\documentclass[lettersize,journal]{IEEEtran}
\usepackage{amsmath,amsfonts}
\usepackage{algorithmic}
\usepackage{algorithm}
\usepackage{array}
\usepackage[caption=false,font=normalsize,labelfont=sf,textfont=sf]{subfig}
\usepackage{textcomp}
\usepackage{stfloats}
\usepackage{url}
\usepackage{verbatim}
\usepackage{graphicx}
\usepackage{cite}
\usepackage{multirow}
\usepackage{tabularx,booktabs}
\newcolumntype{C}{>{\centering\arraybackslash}X} 
\usepackage[dvipsnames]{xcolor}
\usepackage{longtable}
\usepackage{makecell, cellspace, caption}
\usepackage{array}
\usepackage{subcaption}
\usepackage{cuted}
\usepackage{multirow}
\usepackage[utf8]{inputenc}
\usepackage{tabularx}
\usepackage{array}
\usepackage{amssymb}
\usepackage{makecell}
\usepackage[margin=2cm]{geometry}
\usepackage{adjustbox}
\usepackage{multirow}
\usepackage{tabulary,booktabs}
\usepackage{ragged2e}
\usepackage{amsmath}
\usepackage{booktabs}
\usepackage{array}
\newcolumntype{L}{>{}l<{}}
\newcolumntype{C}{>{}c<{}}
\newcolumntype{R}{>{}r<{}}
\newcolumntype{P}{>{}p{3.5em}<{}}
\newcolumntype{F}{>{}p{3.1em}<{}}
\newcolumntype{A}{>{}p{2.5em}<{}}
\newcolumntype{B}{>{}p{3em}<{}}
\newcolumntype{D}{>{}p{2em}<{}}
\newcolumntype{E}{>{}p{1.5em}<{}}
\newcolumntype{G}{>{}p{1em}<{}}

\usepackage{amssymb}
\usepackage{pifont}
\usepackage{graphicx}
\usepackage{hyperref}
\hypersetup{colorlinks,allcolors=black}
\usepackage{flushend}
\usepackage{mathtools}
\usepackage{xurl}
\usepackage{lipsum} 
\usepackage{tikz} 
\begin{document}

\title{LifWavNet: Lifting Wavelet-based Network for Non-contact ECG Reconstruction from Radar}

\author{Soumitra Kundu, Gargi Panda, Saumik Bhattacharya, Aurobinda Routray, Rajlakshmi Guha,~\IEEEmembership{Member,~IEEE}

\thanks{Soumitra Kundu and Rajlakshmi Guha are with the Rekhi Centre of Excellence for the Science of Happiness, IIT Kharagpur, India (e-mail: soumitra2012.kbc@gmail.com; rajg@cet.iitkgp.ac.in).}
\thanks{Gargi Panda and Aurobinda Routray are with the Department of Electrical Engineering, IIT Kharagpur, India
(email: pandagargi@gmail.com; aroutray@ee.iitkgp.ac.in).}
\thanks{Saumik Bhattacharya is with the Department of Electronics and Electrical Communication Engineering, IIT Kharagpur, India
	(email: saumik@ece.iitkgp.ac.in).}}

\markboth{Journal submission}%
{Shell \MakeLowercase{\textit{et al.}}: A Sample Article Using IEEEtran.cls for IEEE Journals}


\maketitle

\begin{abstract}
Non-contact electrocardiogram (ECG) reconstruction from radar signals offers a promising approach for unobtrusive cardiac monitoring. We present LifWavNet, a lifting wavelet network based on a multi-resolution analysis and synthesis (MRAS) model for radar-to-ECG reconstruction. Unlike prior models that use fixed wavelet approaches, LifWavNet employs learnable lifting wavelets with lifting and inverse lifting units to adaptively capture radar signal features and synthesize physiologically meaningful ECG waveforms. To improve reconstruction fidelity, we introduce a multi-resolution short-time Fourier transform (STFT) loss, that enforces consistency with the ground-truth ECG in both temporal and spectral domains. Evaluations on two public datasets demonstrate that LifWavNet outperforms state-of-the-art methods in ECG reconstruction and downstream vital sign estimation (heart rate and heart rate variability). Furthermore, intermediate feature visualization highlights the interpretability of multi-resolution decomposition and synthesis in radar-to-ECG reconstruction. These results establish LifWavNet as a robust framework for radar-based non-contact ECG measurement.
\end{abstract}

\section{Introduction}
\IEEEPARstart{E}{lectrocardiogram} (ECG) is the clinical gold standard for measuring cardiac electrical activity, and its continuous monitoring provides essential information for the diagnosis and management of various cardiac diseases \cite{ecg_golden}. However, conventional ECG systems that use skin-contact electrodes can cause discomfort, skin irritation, and reduced compliance during long-term use—particularly in infants, elderly individuals, and patients with burns or sensitive skin.
Advancements in optical photoplethysmography (PPG) have introduced wearable devices that offer a less obtrusive option, but these still require skin contact. Remote PPG (rPPG) enables contact-free monitoring using visual cameras, yet its performance is constrained by ambient lighting conditions.
Radar sensing has emerged as a promising non-contact alternative, capable of detecting sub-millimeter chest wall displacements caused by cardiac activity, without being affected by lighting conditions. 
Early radar-based methods followed a signal processing pipeline beginning with radar I/Q extraction, demodulation, and phase unwrapping to obtain chest displacement signal. Respiration suppression was then applied using techniques such as bandpass/adaptive filtering, subspace methods, or wavelet decomposition. The cardiac band was subsequently isolated using time–frequency analysis, and heartbeats were tracked through spectral peaks, harmonic summation, or template matching \cite{radar_tutorial,template_tim,cardiac_tim,algorithms_tim,study_tim}.

While these approaches could provide reasonable mean heart rate (HR) estimates, they faced significant limitations when extended to full ECG waveform reconstruction, which requires an accurate representation of all P, Q, R, S, and T waves. Accurate ECG reconstruction from radar remains challenging because the weak cardiac motion signals are often masked by stronger respiratory movements and body motion. Furthermore, the relationship between mechanical and electrical cardiac activity is inherently nonlinear and highly sensitive to noise. In recent years, deep learning (DL) methods \cite{access_doppler, datamobilecomputing,rfecg,radarnet,airecg,autoencoder_sensors,RSSRnet,radarode,radarodemtl,DCGANs,wavegru} have shown considerable promise in addressing these challenges and enabling radar-based ECG reconstruction.

\begin{figure}[t!]
	\centering
	\includegraphics[width=0.49\textwidth]{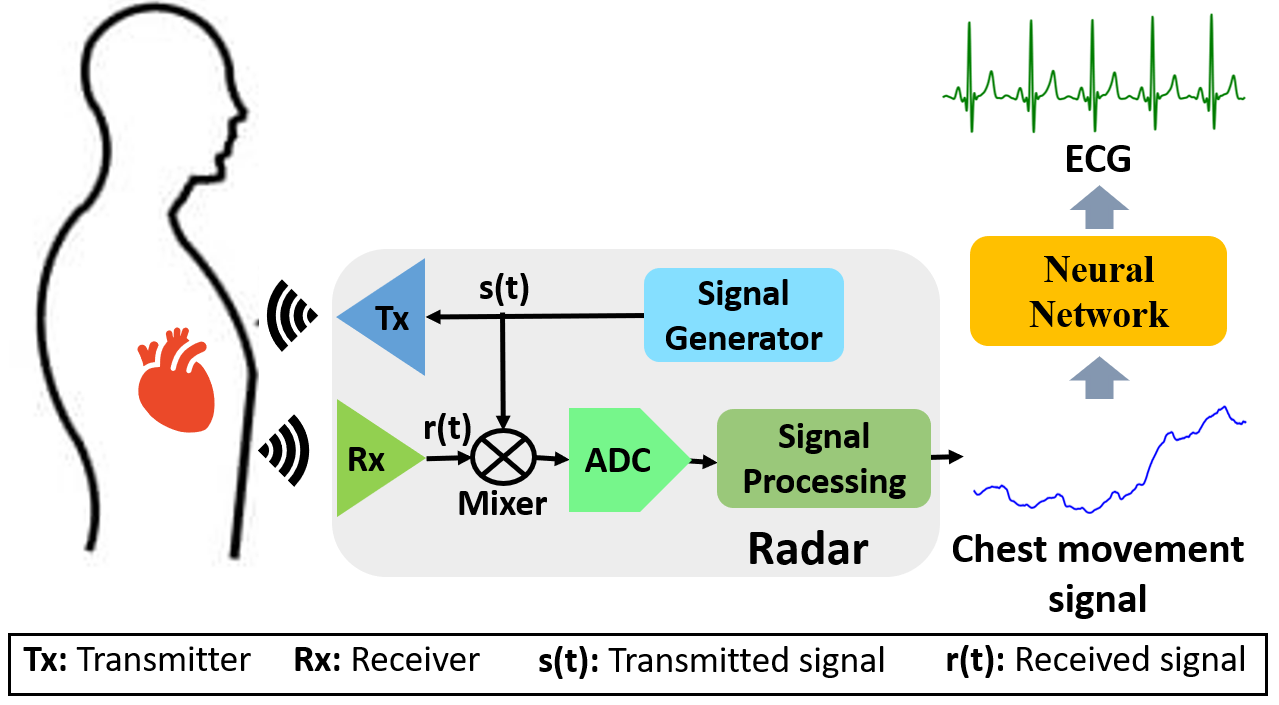}
	\caption{Overview of radar signal-based non-contact ECG reconstruction. The radar measures the chest movement signal in a non-contact manner, from which a neural network reconstructs the ECG signal.}
	\label{fig:concept}
\end{figure}
\subsection{Motivation}
ECG records the heart’s electrical activity through its characteristic PQRST waveform, with each component corresponding to a distinct phase of the cardiac cycle. The P-wave represents atrial depolarization, initiating atrial contraction and blood transfer to the ventricles. The QRS complex, reflecting sequential ventricular depolarization, drives ventricular contraction and blood ejection, while the T-wave denotes ventricular repolarization, leading to relaxation and filling in preparation for the next cycle. This correspondence between the ECG’s electrical events and the heart’s mechanical actions, arising from the excitation–contraction coupling (ECC) mechanism \cite{cardiac_model}, motivates modeling the nonlinear, cross-modal transformation from radar signal to ECG waveform.

Recent deep learning-based radar to ECG reconstruction methods mainly adopt CNN–LSTM \cite{access_doppler}, U-Net \cite{radarnet}, and transformer-based \cite{autoencoder_sensors} architectures, with time and frequency domain-based loss functions to perform the non-linear mapping. Generative models such as GANs \cite{rfecg,DCGANs} and diffusion networks \cite{airecg} employ probabilistic generative perspective to learn the data-distributions. However, these architectures are pure DL-based, designed with trial-and-test strategy, and largely operate as black box. Some recent approaches \cite{datamobilecomputing,radarode,radarodemtl} have proposed to estimate an intermediate 4D cardiac representations to better model the radar to ECG reconstruction, at the cost of heavy pre-processing. Also, the above-mentioned methods lack explicit inductive bias toward the known structure of cardiac signals. This is where wavelets become an important alternative. Unlike Fourier or short time Fourier transform (STFT), which assume fixed resolution across time and frequency, wavelets provide adaptive time–frequency localization. In radar‑to‑ECG, this property is crucial because the radar signal contains overlapping respiratory and cardiac components, and the wavelets provide principled multi-resolution analysis that can isolate sharp transients such as QRS complexes while preserving low frequency morphology (P/T waves and baseline) \cite{wavetheory,ecgwavelet,waveletdetection,embc_wavelet}.

More recently, Xu \textit{et al.} \cite{wavegru} proposed WaveGRUNet, which incorporates Sym4
wavelet based multi-level decomposition as a pre-processing
step, followed by CNN and gated recurrent units (GRUs) for
feature extraction and ECG reconstruction. Similarly, Kim \textit{et al.} \cite{neurofuzzy} integrated wavelet transform with adaptive neuro‑fuzzy networks to preserve nonlinear dependencies while extracting ECG‑relevant frequency bands. Wavelets can effectively encode the multi‑scale nature of cardiac electrophysiology (wideband QRS vs. narrow band P/T). However, prior works have only used fixed, linear wavelet bases (e.g., Sym4, dB4) for pre-processing the radar signal. This underutilizes the potential of wavelets, pointing toward the need for learnable wavelet-based framework that can adaptively capture the nonlinear radar–ECG mapping. Moreover, the existing methods employ a single window STFT-based loss functions to constrain the reconstructed signal with ground truth ECG in both time and frequency domains. The PQRST waveform in ECG signal contain different frequencies, and a multi window-based STFT loss function can be more effective.
Therefore, there is a need for a wavelet-based, learnable framework capable of capturing nonlinear cross-domain mappings. Additionally, incorporating an improved time–frequency constraints seem essential to better preserve the diverse temporal and spectral features of the ECG waveform.

\subsection{Methodology Overview and Contributions}
 Motivated by the insights described above, we first introduce a multi‑resolution analysis and synthesis (MRAS) model for the radar to ECG reconstruction task. Building on this formulation, we design LifWavNet, a novel end‑to‑end network that employs a learnable lifting wavelet scheme. The lifting scheme provides a flexible framework for constructing wavelets, distinct from classical fixed bases such as Haar \cite{lifting_1996}. Unlike these linear, pre‑defined filters, lifting enables the design of nonlinear, learnable filters that can adapt to the underlying structure of the data \cite{dawn}. Prior studies have successfully applied lifting wavelet networks to image classification, compression, and denoising, where they demonstrated improved adaptability and interpretability compared to fixed transforms \cite{dawn,compression1,winnet}. However, these applications involve reconstruction within the same modality. In contrast, radar‑to‑ECG reconstruction is a cross‑domain problem, requiring the transformation of radar echoes into physiologically meaningful ECG waveforms.

In our MRAS model, the radar signal is decomposed into approximation and detail components through multi‑resolution analysis (MRA), and then these components are recombined via multi‑resolution synthesis (MRS) to generate ECG waveforms. LifWavNet is designed based on this model by embedding lifting and inverse lifting units as its core modules. These units are designed to capture essential radar features across scales and synthesize accurate ECG reconstructions, while maintaining interpretability through the wavelet lifting framework. To further improve the reconstruction performance, we propose a multi‑resolution STFT loss to constrain the reconstructed ECG against ground truth in both the temporal and spectral domains. This loss ensures that both transient features (e.g., QRS complexes) and low‑frequency morphology (P/T waves, baseline) are faithfully preserved.
Our contributions can be summarized as follows:
\begin{enumerate}
	\item We propose a novel wavelet-based multi-resolution analysis and synthesis (MRAS) model for the radar to ECG reconstruction task. To the best of our knowledge, this the first work that formulates the radar to ECG reconstruction task using a wavelet-based approach. 
	\item Based on our MRAS model, we design a network, named LifWavNet, for the radar to ECG reconstruction. LifWavNet employs a series of learnable lifting and inverse lifting wavelet filters to perform the MRAS.
	\item We introduce a multi-resolution STFT loss to enforce reconstruction consistency in the time–frequency domain.
	\item We conduct extensive evaluations on two public datasets, showing that our approach achieves superior ECG reconstruction and downstream vital sign estimation (Heart Rate (HR) and Heart Rate Variability
	(HRV)), significantly outperforming state of the art
	methods (SOTA) methods in a unified benchmark. 
\end{enumerate}

The rest of the paper is organized as follows: Section \ref{lit_review} reviews the prior work on radar to ECG reconstruction and the lifting wavelet-based methods. Section \ref{method} presents the proposed methodology. Section \ref{expeiments} reports experimental results. Finally, we conclude the paper in Section \ref{conclusion}.

\section{Background}
\label{lit_review}
\subsection{Prior Work on Radar to ECG Reconstruction}
Recent works on radar-to-ECG reconstruction use deep learning architectures to realize nonlinear, cross-modal mappings between radar and ECG signals. Yamamoto \textit{et al.} \cite{access_doppler} introduced a hybrid CNN–LSTM network capable of detecting R-peaks, but it was less effective in capturing P and T waves. Subsequently, RadarNet \cite{radarnet} presented a U-Net–based architecture with an FFT-based loss function to improve the reconstruction quality. Furthermore, Liu \textit{et al.} \cite{autoencoder_sensors} applied transformer attention to capture long-range dependencies in ECG signals and used an STFT-based loss function to enforce consistency in both time and frequency domains. RSSRNet \cite{RSSRnet} performed STFT on radar signals, carried out ECG reconstruction in the transformed domain, and subsequently applied inverse STFT to recover the final ECG signal.
Moreover, adversarial and generative frameworks have also been explored. RS2ECG \cite{rfecg} leveraged a generative adversarial network (GAN) for radar-to-ECG mapping, while Air-ECG \cite{airecg} employed a diffusion model to better preserve ECG morphology.

Beyond direct mappings, some approaches have introduced intermediate cardiac representations. Chen \textit{et al.} \cite{datamobilecomputing} proposed an algorithm to estimate a 4D cardiac signal from radar echoes as an intermediate representation for ECG reconstruction. Building upon this concept,Zhang \textit{et al.} \cite{radarode} developed RadarODE, an ordinary differential equation (ODE)-embedded network for ECG reconstruction, and later extended it to RadarODE-MTL \cite{radarodemtl}, which incorporated multitask learning (MTL). However, the 4D cardiac signal estimation introduces significant computational overhead during pre-processing.

More recently, Xu \textit{et al.} \cite{wavegru} proposed WaveGRUNet, which applies Sym4 wavelet-based multilevel decomposition to radar signals, extracts sub-band features via CNNs, and reconstructs ECG waveforms using GRUs. Similarly, Kim \textit{et al.} \cite{neurofuzzy} designed an adaptive neuro fuzzy network (ANFN), that extracts features from the multilevel wavelet decomposition of radar signal for the ECG reconstruction. Here, wavelets provide a provable multi-resolution analysis that isolates sharp transients (QRS) while preserving low-frequency morphology (P/T waves and baseline) \cite{wavetheory,ecgwavelet,waveletdetection}. Nevertheless, WaveGRUNet and ANFN employ only a fixed wavelet transform as a pre-processing step. Different from this approach, we propose a wavelet-based MRAS model, based on which we design a network that leverages learnable lifting wavelets for the radar-to-ECG reconstruction task. 
Moreover, the ECG signal has PQRST waveforms, each with a different frequency component. Unlike \cite{autoencoder_sensors,RSSRnet}, that employ single window-based STFT loss function, we propose a multi-resolution STFT loss to better constrain the waveforms in the reconstructed ECG with the ground truth (GT).

\begin{figure}[t!]\captionsetup[subfigure]{font=footnotesize} 
\centering
\subfloat[Wavelet and inverse wavelet transforms using filter banks. ]{\includegraphics[width=0.99\linewidth]{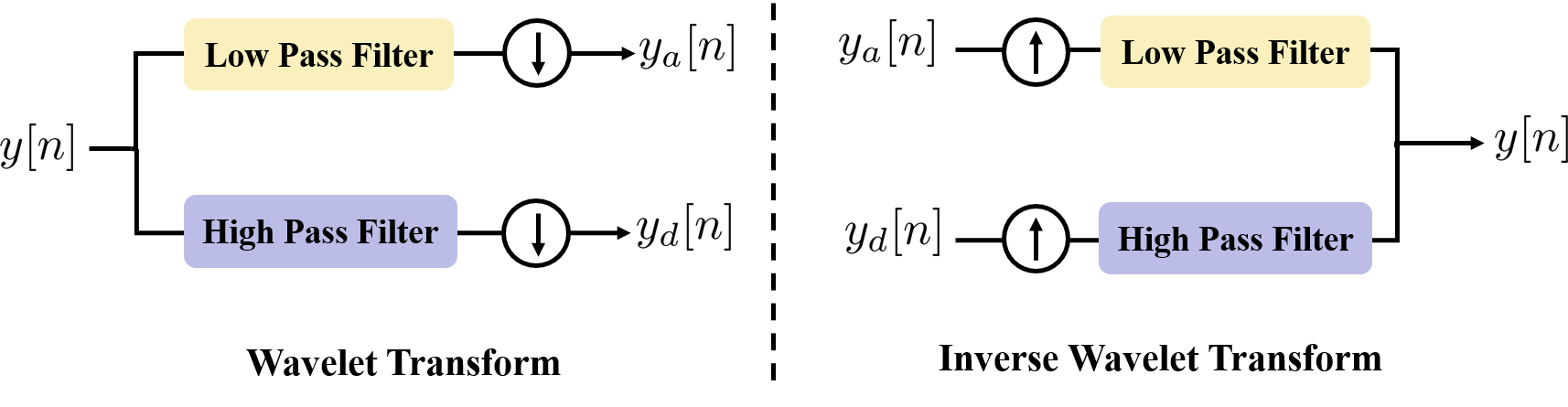}}
\vfill
\subfloat[Wavelet and inverse wavelet transforms using lifting scheme. ]{\includegraphics[width=0.99\linewidth]{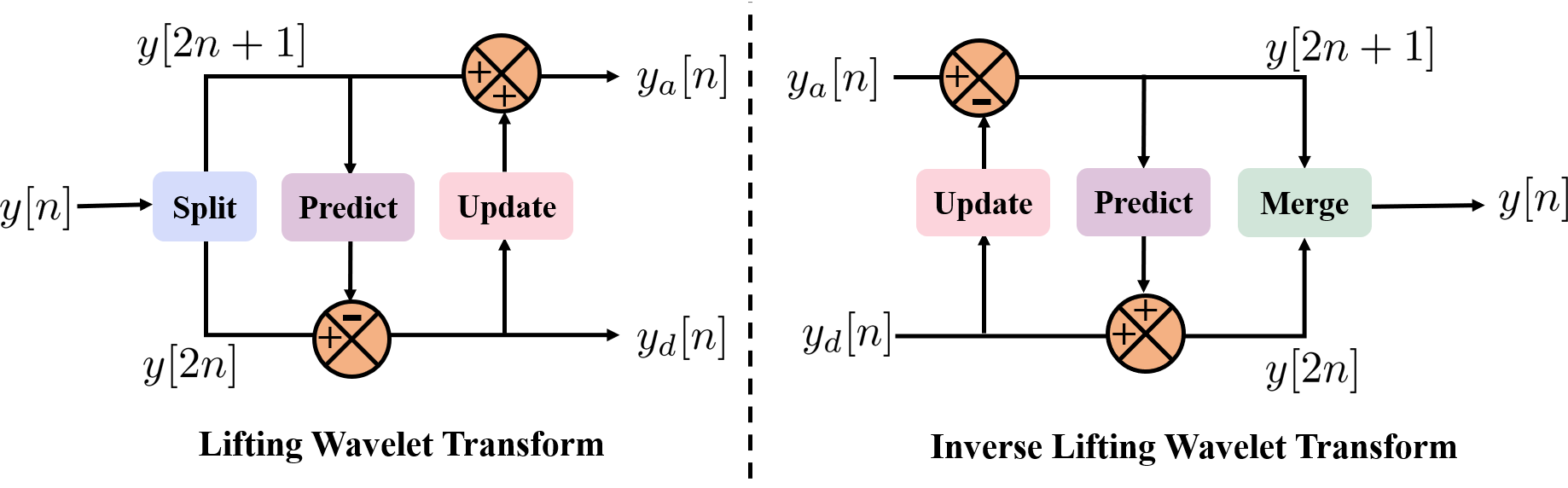}}
\caption{Illustration of wavelet and inverse wavelet transforms using (a) Filter banks, (b) Lifting scheme.}
\label{fig_wavelet} 
\end{figure}
\subsection{Lifting Wavelet Scheme}
Wavelet transform enables signal analysis across multiple frequencies with varying time resolutions and has demonstrated strong performance in both signal and image processing domains \cite{wavetheory,wavelet_denosing,ecgwavelet,signal_classification}. As illustrated in Figure \ref{fig_wavelet}-(a), the wavelet transform decomposes a signal using high-pass and low-pass filters to obtain detailed and approximate components, respectively. The choice of mother wavelets in the filter banks is typically guided by their similarity to the characteristics of the target signal \cite{ecgwavelet}. The inverse wavelet transform can then synthesize the original signal from the decomposed components, ensuring perfect reconstruction. However, classical wavelets such as the Haar wavelet often prove suboptimal, as their limited flexibility restricts their ability to represent complex signals \cite{lifting_1996}. To overcome this limitation, Sweldens \cite{1995lifting} introduced the lifting scheme, a custom-designed method for constructing wavelets. As shown in Figure \ref{fig_wavelet}-(b), both wavelet and inverse wavelet transforms can be realized through this scheme, where the high-pass and low-pass filters are implemented via predict and update steps. The principal advantage of the lifting scheme lies in its ability to construct custom wavelets, which can more effectively capture the underlying correlations in complex signals.

Rodriguez \textit{et al.} \cite{dawn} first introduced a lifting wavelet-based network for image classification, where the transform was implemented with a fixed split step and learnable convolution-based predict and update stages. Similarly, Liu \textit{et al.} \cite{signal_classification} proposed a tree-structured lifting wavelet network for time-series classification. Subsequent studies extended the application of lifting wavelets to image compression \cite{compression1} and image denoising \cite{winnet}. These approaches typically employed convolution-based learnable lifting wavelets that shared parameters across both the analysis and synthesis stages while relying on fixed split and merge operators.

Nevertheless, image compression and denoising tasks reconstruct images within the same domain, whereas radar-to-ECG reconstruction inherently requires a cross-domain transformation. To address this challenge, our proposed LifWavNet employs distinct LUs and ILUs across multiple scales to perform multi-resolution analysis and synthesis. Furthermore, to better capture the nonlinear mapping between radar and ECG signals, our lifting and inverse lifting units are designed with learnable split and merge operators, combined with convolution- and attention-based prediction (P) and updation (U) blocks.

\section{Proposed Method}
\label{method}
\begin{figure*}[hbt!]
	\centering
	\includegraphics[width=0.85\textwidth]{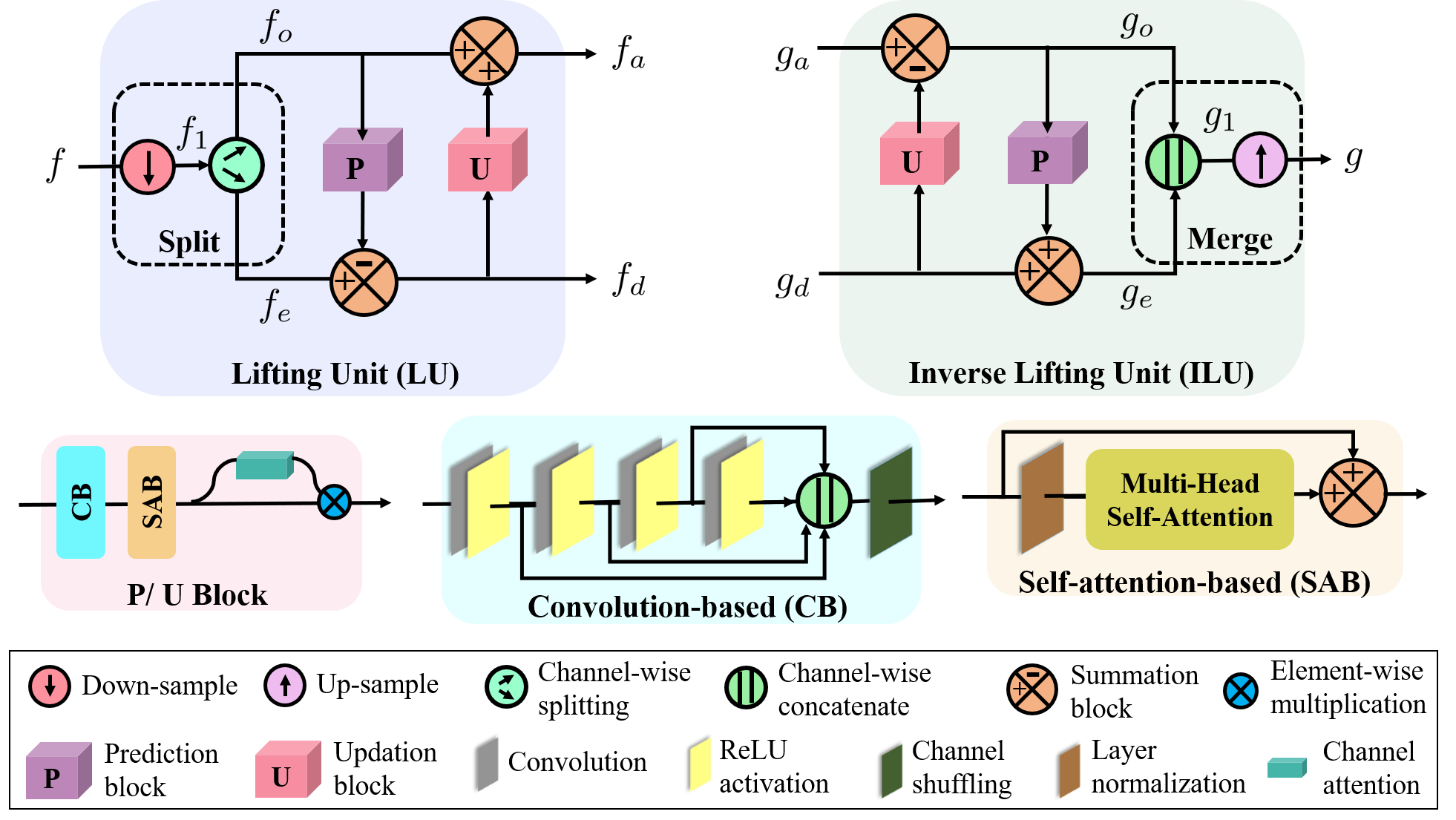}
	\caption{Structure of lifting and inverse lifting units.}
	\label{fig:lus}
\end{figure*}

\subsection{Problem Formulation}
Motivated by the success of wavelet transforms in radar and ECG analysis \cite{ecgwavelet,waveletdetection,wavelet_hr,embc_wavelet}, we formulate the radar‑to‑ECG reconstruction task within a wavelet-based multi‑resolution analysis and synthesis (MRAS) framework. In classical reconstruction task such as image denoising \cite{wavelet_denosing,winnet}, wavelet‑based MRAS decomposes an image into multi‑level approximation and detail coefficients through multi‑resolution analysis (MRA). Then the image is reconstructed back via multi‑resolution synthesis (MRS) using mirrored wavelet filters from the approximation and modified detailed components. This decomposition–synthesis paradigm provides a structured way to capture both coarse and fine features of the image. 
However, unlike image denoising, radar‑to‑ECG reconstruction is not an intra‑domain restoration problem but a cross‑domain transformation: the input radar signal encodes mechanical chest wall motion, while the target ECG represents electrical cardiac activity. This mapping is inherently non-linear and multi‑scale, as sharp QRS complexes coexist with slower P and T waves, and radar echoes are further confounded by respiration and clutter \cite{radar_tutorial}. 

To address this, we extend the MRAS framework to radar‑to‑ECG reconstruction. Given a radar signal $S_R\in \mathbb{R}^{L\times 1}$, we aim to reconstruct the ECG signal $S_E\in \mathbb{R}^{L\times 1}$. Here, $L$ denotes the signal length. We perform multi‑resolution analysis $\text{MRA}(\cdot)$ to decompose $S_R$ into $N$ levels of detail components $d_R^{(1)},d_R^{(2)},...,d_R^{(N)}$ and $N^{th}$ level approximation component $a_R^{(N)}$ as,

\begin{equation}\label{eq_mra_r}
	d_R^{(1)},d_R^{(2)},...,d_R^{(N)},a_R^{(N)} = \text{MRA}(S_R) 
\end{equation}

\noindent These components capture radar features at different temporal and spectral scales. The synthesis stage then recombines them through the multi-resolution synthesis operator $\text{MRS}(\cdot)$ to generate the reconstructed ECG waveform $S_E$ as,

 \begin{equation}\label{eq_mrs_e}
 	S_E  = \text{MRS}\big(d_R^{(1)},d_R^{(2)},...,d_R^{(N)},a_R^{(N)}\big)
 \end{equation}

\noindent Our MRAS formulation provides two key advantages. First, it explicitly separates radar features into scales that correspond to ECG morphology, such as wideband QRS complexes versus narrowband P/T waves. Second, it constrains the reconstruction to follow a principled decomposition–synthesis pathway rather than relying solely on black-box regression, such as pure DL-based methods \cite{access_doppler, rfecg,radarnet,airecg,autoencoder_sensors,DCGANs}.
Building on this formulation, we design LifWavNet, which instantiates MRAS using learnable lifting wavelets. Unlike fixed wavelet bases (e.g., Sym4, dB4), the lifting scheme allows the analysis and synthesis filters to be non-linear and trainable, adapting to radar–ECG nonlinear mappings. We employ a series of lifting and inverse lifting units at different scales, to implement our MRAS model. In the next subsection, we present the design of our lifting unit (LU) and inverse lifting unit (ILU).
\begin{figure*}[hbt!]
	\centering
	\includegraphics[width=\textwidth]{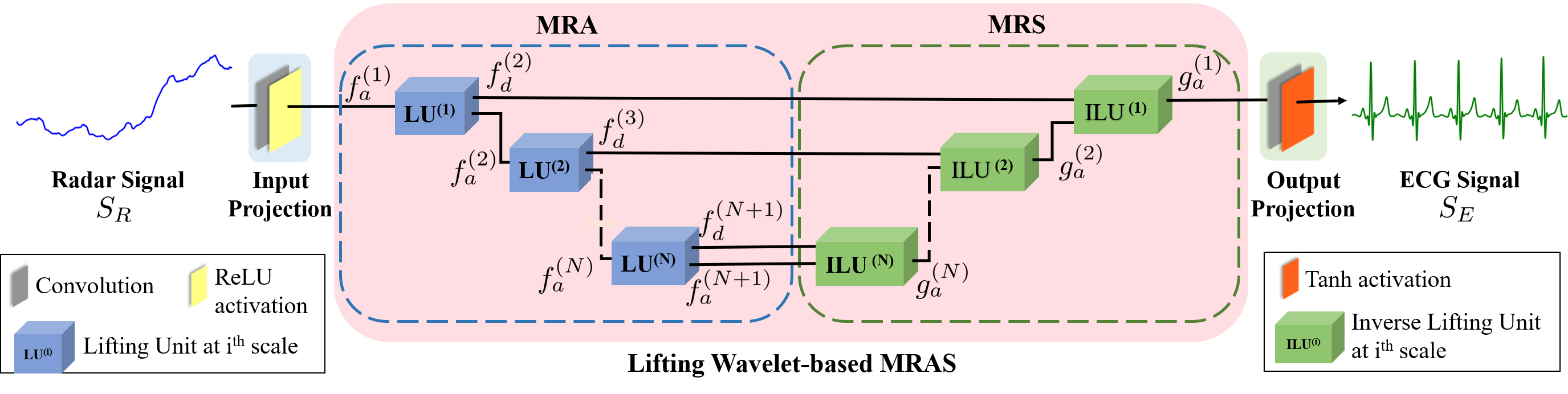}
	\caption{Overall architecture of LifWavNet.}
	\label{fig:archi}
\end{figure*}
\subsection{Lifting and Inverse Lifting Units}
Figure \ref{fig:lus} illustrates the structures of our proposed lifting and inverse lifting units. The lifting unit decomposes the input signal feature $f \in \mathbb{R}^{L\times C}$ into approximation and detail components, $f_a \in \mathbb{R}^{\frac{L}{2}\times C}$ and $f_d \in \mathbb{R}^{\frac{L}{2}\times C}$ respectively. Here, $L$ denotes the signal length, which is even, and $C$ represents the number of channels. Three operations are performed to achieve this decomposition: splitting, prediction, and updation. In the splitting stage, we first use a strided convolution to downsample the signal to $f_1 \in \mathbb{R}^{\frac{L}{2}\times 2C}$. Next, we apply channel-wise splitting to obtain two non-overlapping parts: $f_e \in \mathbb{R}^{\frac{L}{2}\times C}$ and $f_o \in \mathbb{R}^{\frac{L}{2}\times C}$. This design enables a learnable splitting mechanism, instead of manually partitioning the signal into even and odd parts \cite{bp_neural,lifting_reconstruction,dawn}. Following the splitting operation, we apply the prediction ($P(\cdot)$) and updation ($U(\cdot)$) blocks to generate the detailed part $f_d$ and approximation part $f_a$ as,
\begin{equation}\label{eq_c1}
\begin{split}
f_d = f_e-P(f_o) \\
f_a = f_o+U(f_d) 
\end{split}
\end{equation}

\noindent Our prediction and updation blocks are designed with non-linear, learnable operations, whose structures we discuss in the next subsection.

The inverse lifting unit reconstructs the output signal feature $g \in \mathbb{R}^{L\times C}$ from the input approximation and detail components, $g_a \in \mathbb{R}^{\frac{L}{2}\times C}$ and $g_d \in \mathbb{R}^{\frac{L}{2}\times C}$, respectively, by applying three operations: updation, prediction, and merging. Given $g_a$ and $g_d$, we first compute the non-overlapping parts $g_o \in \mathbb{R}^{\frac{L}{2}\times C}$ and $g_e \in \mathbb{R}^{\frac{L}{2}\times C}$ as,
\begin{equation}\label{eq_c1}
\begin{split}
g_o = g_a-U(g_d) \\
g_e = g_d+P(g_o) 
\end{split}
\end{equation}

\noindent where $U(\cdot)$ and $P(\cdot)$ denote the updation and prediction blocks respectively, whose structures are the same as in the lifting unit. Following this, we merge the non-overlapping parts in two stages. First, we perform channel-wise concatenation of $g_o$ and $g_e$ to obtain the feature $g_1 \in \mathbb{R}^{\frac{L}{2}\times 2C}$. Then we apply a strided deconvolution to upsample $g_1$ and produce the reconstructed feature $g \in \mathbb{R}^{L\times C}$. This design provides a learnable merging mechanism, rather than manually combining the even and odd parts to form the signal as in \cite{lifting_reconstruction}. In summary, the lifting unit employs learnable non-linear operations to decompose a signal into detail and approximation components, whereas inverse lifting unit reconstructs the original signal from these decomposed components. In the next subsection, we explain the structures of prediction and updation blocks, which form the core modules of LU and ILU.
\subsection{Structure of Prediction and Updation Blocks}
The same network structure has been used for the prediction (P) and updation (U) blocks, as shown in Figure \ref{fig:lus}. These blocks together approximate the filters of wavelet transforms. By employing an end-to-end training strategy, the parameters of these blocks are learnt from the data. Furthermore, the blocks utilize non-linear, input-dependent operations, which facilitate efficient learning of the wavelet filter structures.

We use convolution and attention mechanisms to design the  P and U blocks. First, a convolution-based layer is employed to model the local structural details of the signal. Following this, a self-attention-based  layer is applied as an input-dependent non-linear operation. Finally, we incorporate a channel-attention layer \cite{channel_attention}, to capture more distinctive information from the features along the channel dimension.

In the convolution-based layer, we apply a series of convolution and ReLU operations to capture the local structural details of the input signal. We adopt the CSConv layer \cite{latis}, where four convolution operations with different kernel sizes are applied sequentially. This design enables the extraction of features at multiple receptive fields. The resulting features are then concatenated along the channel dimension, followed by channel-shuffling, where the feature channels are shuffled to promote information exchange  among the aggregated features. 

Although convolution is highly effective at extracting local structures from the signal, it remains a linear operation. The purpose of the self-attention-based layer is to introduce an input-dependent non-linear transformation on the feature representation. First, layer normalization \cite{layerNorm} is applied to normalize the distribution of intermediate features, followed by a multi-head self attention operation. Here, we adopt a 1-D variant of the self-attention layer proposed in \cite{vit}. By deriving queries, keys, and values from the input feature, the self-attention layer functions as an input-dependent non-linear operator. This non-linearity enables better modeling of the underlying filters of the wavelet transform.

The prediction and updation blocks act as the core modules of our lifting and inverse lifting units. In the next subsection, we describe how these units are integrated into the overall architecture of our network.

\subsection{Overall architecture of LifWavNet}
Given the input radar signal $S_R\in\mathbb{R}^{L\times 1}$, we aim to generate the corresponding ECG signal $S_E\in\mathbb{R}^{L\times 1}$. Figure \ref{fig:archi} illustrates the overall architecture of LifWavNet.  The input radar signal is processed in three main steps: input projection, lifting wavelet-based multi-scale reconstruction, and output projection. In the input projection step, we apply a convolution layer to $S_R$ to get the input feature $f_a^{(1)}\in\mathbb{R}^{L\times C}$, where $C$ denotes the number of feature channels. We do MRAS on this embedded feature, instead of the raw signal. In MRAS, $f^{(1)}$ undergoes a series of lifting and inverse lifting units at different scales to produce the reconstructed feature 	
$g_a^{(1)}\in\mathbb{R}^{L\times C}$.	
In the output projection step, a final convolution layer is applied to $g^{(1)}$ to generate the output ECG signal $S_E\in\mathbb{R}^{L\times 1}$.

The lifting wavelet-based MRAS stage performs the analysis and synthesis of the feature $f_a^{(1)}$ across $N$ different scales. At each scale $i$, there is the lifting unit $\text{LU}^{(i)}$ and inverse lifting unit $\text{ILU}^{(i)}$ to perform analysis and synthesis, respectively. At the $i^{th}$ scale, the feature $f_a^{(i)}\in\mathbb{R}^{\frac{L}{2^{i-1}}\times C}$ is input to $\text{LU}^{(i)}$, which decomposes it into an approximation component $f_a^{(i+1)}\in\mathbb{R}^{\frac{L}{2^{i}}\times C}$ and a detail component $f_d^{(i+1)}\in\mathbb{R}^{\frac{L}{2^{i}}\times C}$. This analysis process can be formulated as,

\begin{equation} \label{eq_lu}
	f_a^{(i+1)},f_d^{(i+1)} = \text{LU}^{(i)}\Big(f_a^{(i)}\Big) 
\end{equation}

\noindent The approximation component is passed to the $\text{LU}^{(i+1)}$, and the detail component is fed into $\text{ILU}^{(i)}$. For the inverse lifting unit $\text{ILU}^{(i)}$ at the $i^{th}$ scale, the feature $g_a^{(i)}\in\mathbb{R}^{\frac{L}{2^{i-1}}\times C}$ is synthesized by taking the approximate feature $g_a^{(i+1)}\in\mathbb{R}^{\frac{L}{2^{i}}\times C}$ from the $\text{ILU}^{(i+1)}$, and the detail feature $f_d^{(i+1)}\in\mathbb{R}^{\frac{L}{2^{i}}\times C}$ from the $\text{LU}^{(i)}$ as input. This synthesis process can be formulated as,

\begin{equation} \label{eq_ilu}
	g_a^{(i)} = \text{ILU}^{(i)}\Big(g_a^{(i+1)},f_d^{(i+1)}\Big)  
\end{equation}

\noindent At $N^{th}$ scale, the outputs from $\text{LU}^{(N)}$ are fed into $\text{ILU}^{(N)}$ to obtain the reconstructed feature $g_a^{(N)}\in\mathbb{R}^{\frac{L}{2^{N-1}}\times C}$.  This multi-resolution analysis and synthesis is performed to reconstruct the essential ECG feature representation from the radar signal.

We design the structures of the prediction and updation blocks for the LUs and ILUs to be identical across different scales. However, their parameters are not shared and the LUs and ILUs at each scale learn their own set of parameters during training. This design choice enables a more flexible adaption of the underlying wavelet filters, thereby improving the analysis and synthesis of features at different scales.
\begin{figure}[t!]
	\centering
	\includegraphics[width=0.5\textwidth]{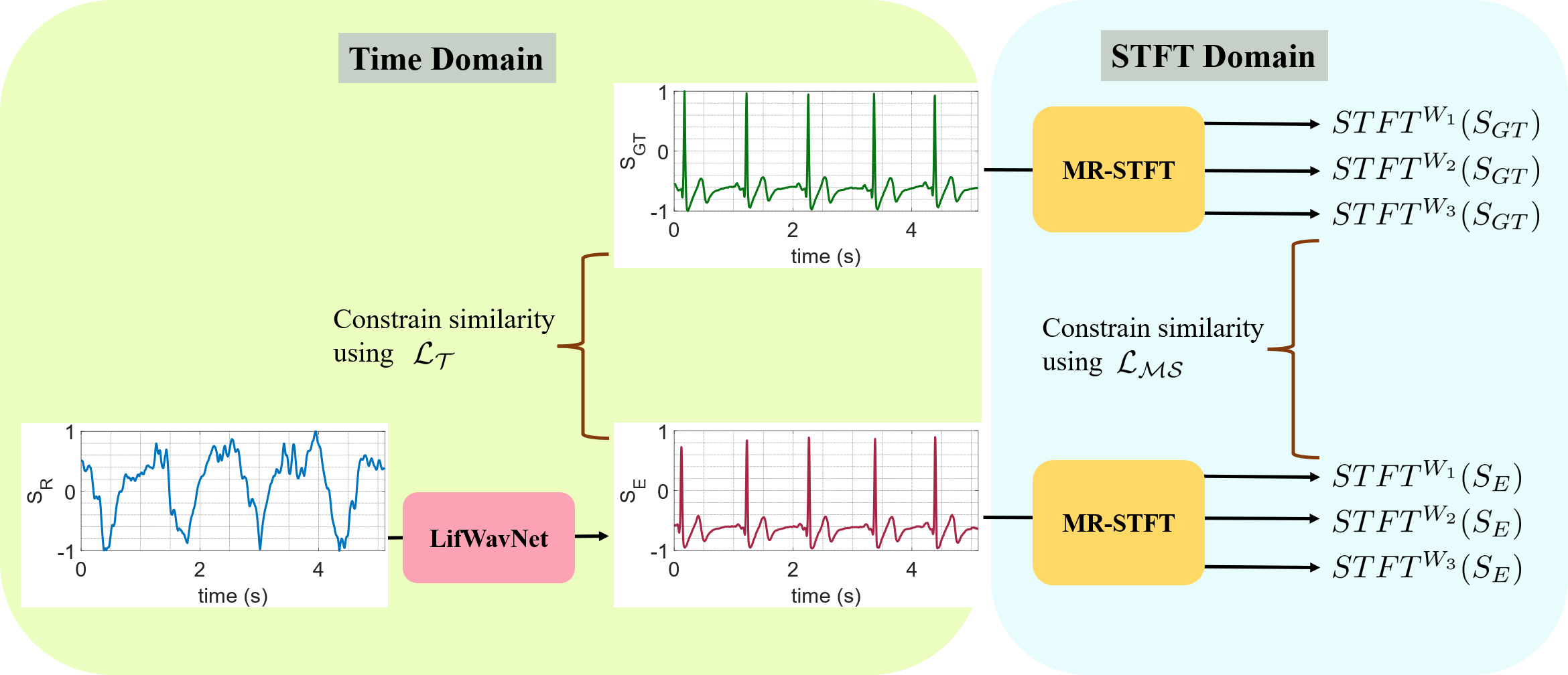}
	\caption{Our loss function constrains the similarity between reconstructed ($S_E$) and ground-truth ECG signal ($S_{GT}$) in both time and STFT domains. The multi-resolution STFT (MR-STFT) performs STFT with different window lengths.}
	\label{fig:spectrum}
\end{figure}
\subsection{Loss Function}
During the training of LifWavNet, the generated ECG signal $S_E$ is constrained to be similar with the ground-truth ECG signal $S_{GT}$. We constrain the time and STFT domain similarities between $S_E$ and $S_{GT}$. To maintain similarity in time domain, our temporal loss function ($\mathcal{L_{T}}$) uses the $L_1$ norm of the difference between the two signals, defined as,

\begin{equation}\label{eq_time}
	\mathcal{L_{T}}=||S_E-S_{GT}||_1
\end{equation}

\noindent The STFT analyzes how the frequency content of a non-stationary signal changes over time. The STFT of a signal is computed by sliding an analysis window of length $W$ over the signal and calculating the discrete Fourier transform (DFT) of each segment of windowed data. The window hops over the signal, and the DFT of each segment is added to a complex-valued matrix that contains the magnitude and phase of each point in time and frequency. In the STFT-based time-frequency representation of signals, there exists a trade-off between time and frequency resolutions - increasing window size gives higher frequency resolution while reducing temporal resolution, and vice versa. By combining multiple STFT losses with different window sizes, the network can better learn the different time-frequency characteristics of the ECG signal. Our multi-resolution STFT loss is the average of the STFT losses computed with three different window sizes. If $STFT^{W_i}(\cdot)$ denotes the STFT operation with the $i^{th}$ window size $W_i$, then our MR-STFT loss is defined as,

\begin{equation}\label{eq_MS}
	\mathcal{L_{MS}}=\frac{1}{3}\sum_{i=1}^{3}||STFT^{W_i}(S_E)-STFT^{W_i}(S_{GT})||_1
\end{equation}

\noindent Our overall loss function becomes,
\begin{equation}\label{eq_loss}
		\mathcal{L}= \mathcal{L}_T+\alpha\mathcal{L}_{MS}
\end{equation}

\noindent Empirically we set the weight $\alpha$ to $0.1$.

\section{Experiments}
\label{expeiments}
We have conducted extensive experiments to evaluate the performance of our proposed method. In Section \ref{exp_setup}, we describe the experimental setup, detailing the datasets, implementation and training settings, and evaluation metrics. Section \ref{exp_sota} presents the quantitative and qualitative performance of LifWavNet compared with the SOTA methods. Then in Section \ref{features}, we visualize the intermediate features of LifWavNet, which provides good interpretability of our method. Finally, in Section \ref{exp_ablation}, we conduct an ablation study to validate the design choices in our proposed network and the loss function.

\subsection{Setup}
\label{exp_setup}
\noindent \textbf{Datasets: } We used two publicly available datasets: Clinically Recorded Radar Vital Signs (CR-RVS) \cite{nature} and medical radar (Med-Radar) \cite{elsevier}, for training and testing our proposed method. These two datasets are described below.

\noindent i) CR-RVS: This dataset is a collection of clinically recorded vital signs by Schellenberger \textit{et al.} \cite{nature}, featuring synchronized radar and multi-modal physiological reference signals. The data were acquired from 30 healthy subjects (14 males and 16 females) in a clinical setting at the University Hospital Erlangen. During the experiments, subjects were positioned on a tilt table while a 24 GHz continuous-wave (CW) six-port radar system, located approximately 40 cm away, was aimed at their chest to capture cardio-respiratory-induced surface motion. The ground-truth electrocardiogram (ECG) was recorded using a medical-grade Task Force Monitor (TFM). During the experiments, subjects were asked to maintain a quasi-static posture to ensure a good signal-to-noise ratio (SNR) with minimal random body movement (RBM) noise. The radar and TFM systems recorded data asynchronously, with sampling rates of 2000 Hz and 1000 Hz, respectively. A sophisticated post-processing synchronization procedure was employed, utilizing a shared Gold code sequence and cross-correlation to precisely correct the time lag between the two systems. Furthermore, an optimization routine determined an exact resampling factor to compensate for any clock drift, ensuring sub-sample alignment between the radar displacement signal and the reference ECG signal. 
 
\noindent ii) Med-Radar: This dataset, collected by Edanami and Sun \cite{elsevier}, contains recordings from nine healthy subjects (five males and four females) in a laboratory environment. The subjects were in a resting state, lying on a bed for 10-minute recordings. Non-contact measurements were conducted using a 24.25 GHz Doppler radar unit with I/Q channels, placed approximately 15 cm beneath the bed. Synchronous ground-truth data were acquired using a BIOPAC system, which provided a reference ECG signal. All signals in this dataset were synchronously sampled at a rate of 1000 Hz.

For both datasets, the radar I/Q signals were first processed to derive the chest-wall displacement time-series signal following \cite{neonatal}, which serves as the input to our reconstruction network, while the corresponding, precisely synchronized ECG recordings from the respective reference devices provide the ground truth. For both datasets, we resampled the synchronized radar and ECG signals to a sampling frequency of 200 Hz. We then split the datasets into training and test sets. The signals were segmented into 5.12 s chunks without overlap, with each segment having a length of 1024 samples. For the CR-RVS dataset, we used 1,292 signal chunks for training and 70 for testing. For the Med-Radar dataset, we used 1,000 signal chunks for training and 53 for testing. As a pre-processing step, the signal chunks were scaled to the range $[-1,1]$.

\noindent \textbf{Implementation details: } We set the number of scales $N$ to $4$ in LifWavNet. The four convolution layers in CSConv have $8$ filters each, with kernel sizes of $31\times 1$, $33\times 1$, $35\times 1$, and $37\times 1$. Other convolutional layers in LifWavNet have a kernel size of $31\times 1$, and number of filters $C=32$. For our MR-STFT loss in Eqn. \ref{eq_MS}, we use Hanning window with three window lengths $W_1,W_2,W_3$ of $800$, $400$, and $200$, respectively.

\noindent \textbf{Training settings: } We train LifWavNet by minimizing the loss function in Eqn. \ref{eq_loss}. Training is performed using the Adam optimizer for $1000$ epochs, with a batch size of $256$, and a learning rate of $1\times 10^{-4}$.

\noindent \textbf{Evaluation metrics: } We use the Pearson correlation coefficient ($\rho$) and the mean relative error (MRE) to evaluate the accuracy of the reconstructed ECG $S_E$ against the ground truth signal $S_{GT}$. The metrics are defined as,

\begin{equation}
	\rho = \frac{ (S_{GT} - \mu[S_{GT}])^T(S_E - \mu[S_E])}{ ||S_{GT} - \mu[S_{GT}]||_2||S_E - \mu[S_E]||_2}
\end{equation}

\begin{equation}
	\text{MRE} = \frac{ ||S_{GT} -S_E||_1}{ ||S_{GT}||_1}
\end{equation}

\noindent where $\mu[\cdot]$ and $||\cdot||_2$ denote the element-wise mean and $L_2$ norm, respectively. A higher value of $\rho$ and a lower value of MRE indicate better signal reconstruction. 

\begin{figure}[t!]
	\centering
	\includegraphics[width=0.45\textwidth]{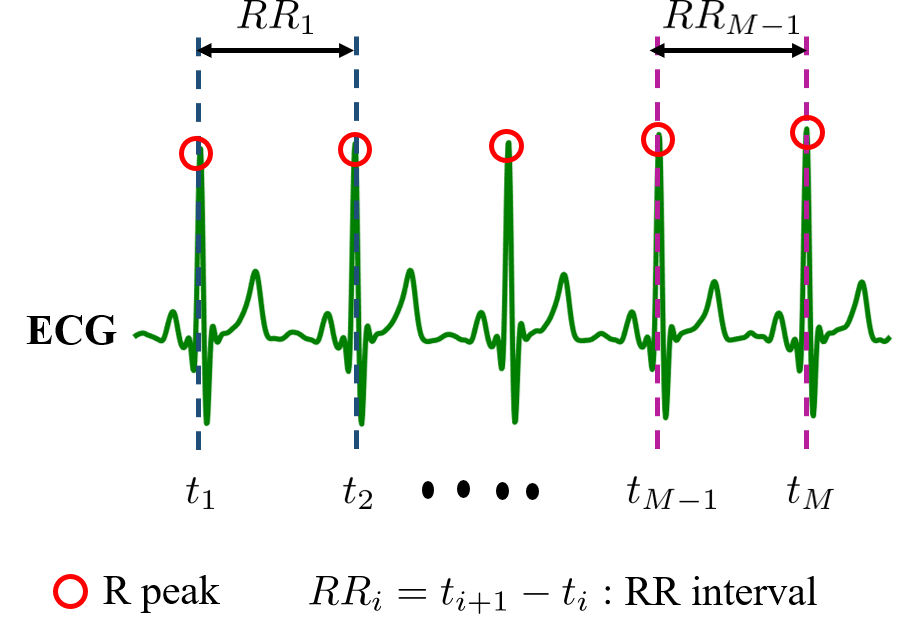}
	\caption{Illustration of detected R peaks and the RR intervals in an ECG signal.}
	\label{fig:rr}
\end{figure}

Moreover, we also estimate the HR and HRV from the reconstructed ECG, and compare these with the corresponding values from the ground-truth ECG. We use the BioSPPy toolbox \footnote[1]{Available online: \url{https://biosppy.readthedocs.io/en/stable/\#\%23welcome-to-biosppy}} to detect the R-peaks in the ECG signal. Figure \ref{fig:rr} illustrates the detected R peaks and the RR intervals in an ECG signal. The HR is calculated as,
\begin{equation}
	\text{HR} = \frac{M-1}{\sum_{i=1}^{M - 1} RR_i} 
\end{equation}

\noindent where $M$ is the number of R peaks detected. Then, in order to estimate the HRV parameter, we calculate the root mean square of successive differences (RMSSD), defined as:
\begin{equation}
	\text{RMSSD} = \sqrt{\frac{\sum_{i=1}^{M - 2} (RR_{i+1}-RR_i)^2}{M-2}} 
\end{equation}

\noindent We compare the estimated HR and RMSSD from the reconstructed ECG with those from the ground truth ECG using the mean absolute error (MAE). The MAE for HR is defined as,
\begin{equation}
	\text{MAE}_{\text{HR}} = ||\text{HR}_{S_{GT}} -\text{HR}_{S_E}||_1
\end{equation}

\noindent The MAE for RMSSD is defined as,

\begin{equation}
	\text{MAE}_{\text{RMSSD}} = ||\text{RMSSD}_{S_{GT}} -\text{RMSSD}_{S_E}||_1
\end{equation}

\noindent For better estimation of vitals, lower values of $\text{MAE}_{\text{HR}}$ and $\text{MAE}_{\text{RMSSD}}$ are desired. 
\begin{table*} 
\fontsize{7.3}{7.7}\selectfont
\centering
\caption{Performance comparison with the SOTA methods on ECG reconstruction from radar signal. \textit{We train all the methods with the same training dataset before testing}. FLOPs and runtime are calculated under the setting of reconstructing an ECG signal from input radar signal of length $1024$. The $\text{MAE}_{\text{HR}}$ and $\text{MAE}_{\text{RMSSD}}$ are measured in bpm and ms units, respectively. The best and second-best performances are highlighted in \textcolor{red}{\textbf{red}} and \textcolor{blue}{\textbf{blue}} colors, respectively.  $\downarrow$ means low value desired, and $\uparrow$ means high value desired.}
\begin{tabular}{R|CCC|CCCC|CCCC}
\toprule
\multirow{3}{*}{Method} &\multirow{3}{*}{\makecell{Params\\(K)}} & \multirow{3}{*}{\makecell{FLOPs\\(G)}} &\multirow{3}{*}{\makecell{Runtime\\(sec)}} &
\multicolumn{4}{c|}{CR-RVS \cite{nature}} &
\multicolumn{4}{c}{Med-Radar \cite{elsevier}} \\
\cmidrule(lr){5-8}
\cmidrule(lr){9-12}

&&&&
\multicolumn{2}{c}{Signal Reconstruction}     &
\multicolumn{2}{c|}{Vitals Estimation}     &
\multicolumn{2}{c}{Signal Reconstruction}     &
\multicolumn{2}{c}{Vitals Estimation} \\
\cmidrule(lr){5-6}
\cmidrule(lr){7-8}
\cmidrule(lr){9-10}
\cmidrule(lr){11-12}

&&&&
\multicolumn{1}{c}{$\rho\uparrow$} &
\multicolumn{1}{c}{MRE$\downarrow$}     &
\multicolumn{1}{c}{$\text{MAE}_{\text{HR}}\downarrow$} &
\multicolumn{1}{c|}{$\text{MAE}_{\text{RMSSD}}\downarrow$}     &
\multicolumn{1}{c}{$\rho\uparrow$} &
\multicolumn{1}{c}{MRE$\downarrow$}     &
\multicolumn{1}{c}{$\text{MAE}_{\text{HR}}\downarrow$} &
\multicolumn{1}{c}{$\text{MAE}_{\text{RMSSD}}\downarrow$}\\
\midrule
\text{RF2ESG \cite{rfecg}}  &233
& 0.239 & 0.003& \textcolor[rgb]{ 0,  0,  1}{\textbf{0.863}} & \textcolor[rgb]{ 0,  0,  1}{\textbf{1.593}} & \textcolor[rgb]{ 0,  0,  1}{\textbf{2.55}} & \textcolor[rgb]{ 0,  0,  1}{\textbf{44.74}} & \textcolor[rgb]{ 0,  0,  1}{\textbf{0.212}} & \textcolor[rgb]{ 0,  0,  1}{\textbf{0.909}} & 24.46 & 493.58 \\

\midrule
\text{AirECG \cite{airecg}}  &20052 & 2.727 & 8.618& 0.000 & 4.217 & 8.14  & 210.56 & 0.012 & 1.697 & \textcolor[rgb]{ 0,  0,  1}{\textbf{12.89}} & \textcolor[rgb]{ 0,  0,  1}{\textbf{232.36}} \\

\midrule
\text{Ours} &990 & 0.251 & 0.032& \textcolor[rgb]{ 1,  0,  0}{\textbf{0.926}} & \textcolor[rgb]{ 1,  0,  0}{\textbf{0.823}} & \textcolor[rgb]{ 1,  0,  0}{\textbf{0.63}} & \textcolor[rgb]{ 1,  0,  0}{\textbf{13.67}} & \textcolor[rgb]{ 1,  0,  0}{\textbf{0.723}} & \textcolor[rgb]{ 1,  0,  0}{\textbf{0.451}} & \textcolor[rgb]{ 1,  0,  0}{\textbf{0.52}} & \textcolor[rgb]{ 1,  0,  0}{\textbf{12.29}} \\

\bottomrule
\end{tabular}
\label{tab:sota_main}
\end{table*}
\begin{figure*}[t!]
	\centering
	\includegraphics[width=\textwidth]{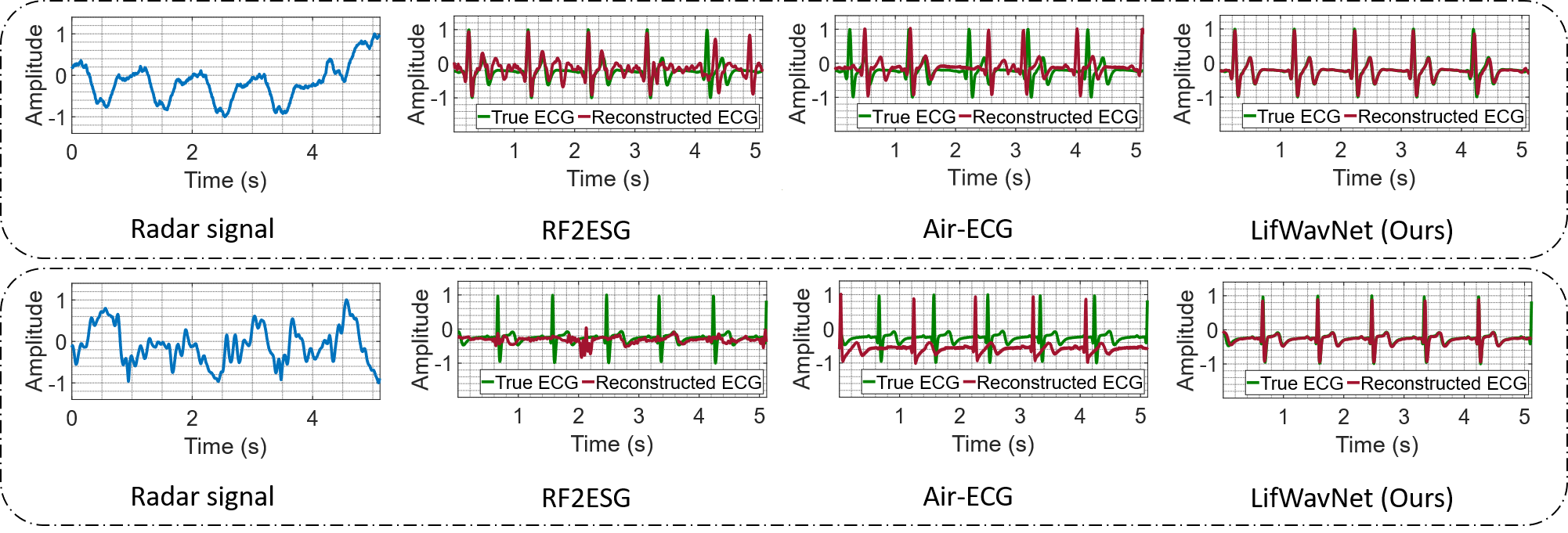}
	\caption{Visual comparison with SOTA methods. The first row shows a signal from the CR-RVS dataset \cite{nature}, and the second row shows a signal from the Med-Radar dataset \cite{elsevier}. Signals are best viewed in $200\%$ zoom.}
	\label{fig:sota}
\end{figure*}
\subsection{Performance Comparison with SOTA Methods}
\label{exp_sota}
We compare our proposed LifWavNet with SOTA radar‑to‑ECG reconstruction methods that operate directly on radar time series data. Among existing approaches, only RF2ESG \footnote[2]{Github code: \url{https://github.com/wzhaha/UWB2ECG
}} 
\cite{rfecg} and Air-ECG \footnote[3]{Github code: \url{https://github.com/LangchengZhao/AirECG}} 
\cite{airecg} provide open‑source code; therefore, we restrict our comparison to these methods. To ensure fairness, all models were trained on the same datasets, and for the SOTA methods we adopted the hyperparameter settings provided in their official code.

\noindent \textbf{Quantitative comparison: }
Table \ref{tab:sota_main} reports the  reconstruction and vitals estimation performance across the CR‑RVS and Med‑Radar datasets. LifWavNet consistently achieves the best results across all metrics. In particular, it yields higher correlation coefficients $(\rho)$ and lower mean relative error (MRE) and mean absolute errors (MAE) compared to RF2ESG and Air‑ECG. These improvements demonstrate that the proposed MRAS‑based design with learnable lifting wavelets provides more accurate ECG waveform reconstruction and more reliable downstream HR/HRV estimation. 

We also compare the computational complexity of LifWavNet with the SOTA methods. RF2ESG is lightweight but less accurate, while Air‑ECG is computationally expensive, requiring over $20$M parameters and long inference times. LifWavNet achieves a favorable balance: although it uses more parameters than RF2ESG, it achieves superior accuracy. Its runtime remains practical for real‑time applications, highlighting the efficiency of the learnable lifting wavelet design.

\noindent \textbf{Qualitative comparison: }
Figure \ref{fig:sota} presents a visual comparison of reconstructed ECG signals. For both datasets, RF2ESG and Air‑ECG fail to capture fine‑scale morphology, often distorting P/T waves or misaligning QRS waveforms. In contrast, reconstructed ECG from LifWavNet closely follow the ground‑truth ECG, preserving both sharp transients and low‑frequency components. This qualitative evidence reinforces the quantitative results reported in Table \ref{tab:sota_main}.
\begin{figure}
	\centering
	\includegraphics[width=0.45\textwidth]{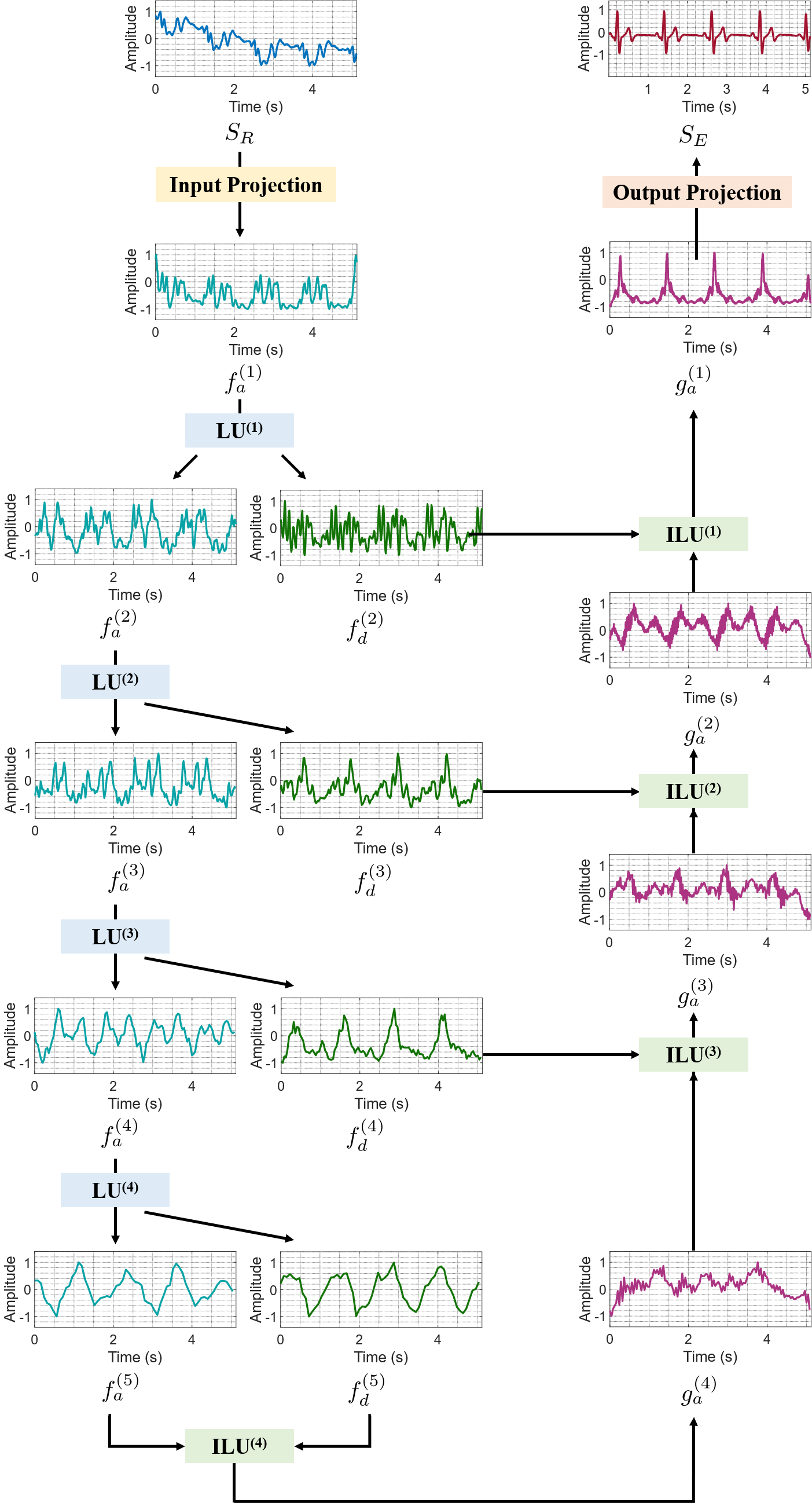}
	\caption{Visualization of intermediate features for a radar signal from CR-RVS dataset. }
	\label{fig:features}
\end{figure}
\subsection{Visualization of Intermediate Features}
\label{features}
Our proposed LifWavNet is designed around the MRAS principle, where the radar signal is first decomposed into multi‑scale approximation and detail components, and these components are subsequently recombined to synthesize the ECG waveform. To better understand the internal working of our network, we visualize in Figure \ref{fig:features} the intermediate features obtained for a representative radar signal from the CR‑RVS dataset. 

The figure illustrates both input projections (features extracted during the analysis stage) and output projections (features synthesized during the reconstruction stage) across multiple decomposition levels. At lower levels ($\text{LU}^{(1)}$,$\text{LU}^{(2)}$), the features retain coarse temporal structure, dominated by respiratory motion. As the decomposition depth increases ($\text{LU}^{(3)}$,$\text{LU}^{(4)}$), the network progressively isolates finer‑scale oscillations corresponding to cardiac activity. This hierarchical separation demonstrates how LifWavNet disentangles overlapping physiological components: the approximation coefficients capture low‑frequency morphology, while the detail coefficients emphasize sharp transients such as QRS complexes.

Importantly, the comparison between input and output projections highlights the synthesis capability of the network. While the input projections contain mixed and noisy radar features, the output projections show progressively refined structures that align more closely with ECG morphology. This validates the design of LifWavNet’s lifting and inverse‑lifting units, which not only decompose radar signals into physiologically meaningful scales but also recombine them into coherent ECG reconstructions.

Thus, the visualization provides interpretability to the MRAS framework: it confirms that LifWavNet leverages multi‑resolution decomposition to separate cardiac information from radar echoes and then synthesizes these features into accurate ECG waveforms.
\subsection{Ablation Study}
\label{exp_ablation}
The performance of LifWavNet is dependent on both its architectural design and the choice of loss function. In particular, the network leverages MRAS with learnable lifting wavelet filters to extract essential radar features for ECG reconstruction, while our loss function constrains the reconstructed ECG against ground truth in both time and frequency domains. To validate these design choices, we conduct a series of ablation experiments on the Med‑Radar dataset, reporting results for both signal reconstruction (SR) and vitals estimation (VE).

\noindent \textbf{Effect of changing the number of scales: } Table \ref{tab:ablation_design1} shows the impact of varying the number of decomposition scales in LifWavNet. Using four scales yields the best overall performance, achieving higher correlation $(\rho)$ and lower errors in HR and HRV estimation. Decreasing the number of scales to three or increasing the number of scales to five degrade performance. 

\begin{table}[t!]
\centering
\caption{Effect of changing the number of scales.}
\begin{tabular}{ccccc}
\toprule
\multicolumn{2}{c}{No. of scales} & \text{3}     & \text{4}     & \text{5} \\
\midrule
\multicolumn{1}{c}{\multirow{2}[2]{*}{SR}} & \multicolumn{1}{c}{$\rho \uparrow$} & \multicolumn{1}{c}{\textcolor[rgb]{ 0,  0,  1}{\textbf{0.721}}} & \textcolor[rgb]{ 1,  0,  0}{\textbf{0.723}} & \multicolumn{1}{c}{0.568} \\

\cmidrule{2-5}          & \multicolumn{1}{c}{MRE $\downarrow$} & \multicolumn{1}{c}{\textcolor[rgb]{ 1,  0,  0}{\textbf{0.381}}} & \textcolor[rgb]{ 0,  0,  1}{\textbf{0.451}} & \multicolumn{1}{c}{0.637} \\

\cmidrule{1-5}    \multicolumn{1}{c}{\multirow{2}[2]{*}{VE}} & \multicolumn{1}{c}{$\text{MAE}_{\text{HR}}$ (bpm) $\downarrow$} & \multicolumn{1}{c}{\textcolor[rgb]{ 0,  0,  1}{\textbf{1.31}}} & \textcolor[rgb]{ 1,  0,  0}{\textbf{0.52}} & \multicolumn{1}{c}{1.70} \\

\cmidrule{2-5}          & $\text{MAE}_{\text{RMSSD}}$ (ms) $\downarrow$ & \text{72.85} & \textcolor[rgb]{ 1,  0,  0}{\textbf{12.29}} & \textcolor[rgb]{ 0,  0,  1}{\textbf{46.64}} \\

\bottomrule    \end{tabular}
\label{tab:ablation_design1}
\vspace*{1em}

\caption{Effect of parameter sharing across scales.}
\begin{tabular}{ccccc}
	\toprule
	\multicolumn{2}{c}{Configuration} & NC-I     & NC-II     & LifWavNet \\
	\midrule
	\multicolumn{1}{c}{\multirow{2}[2]{*}{SR}} & \multicolumn{1}{c}{$\rho \uparrow$} & 0.715 & \textcolor[rgb]{ 0,  0,  1}{\textbf{0.721}} & \textcolor[rgb]{ 1,  0,  0}{\textbf{0.723}} \\
	
	\cmidrule{2-5}          & \multicolumn{1}{c}{MRE $\downarrow$} & 0.527 & \textcolor[rgb]{ 1,  0,  0}{\textbf{0.409}} & \textcolor[rgb]{ 0,  0,  1}{\textbf{0.451}} \\
	
	\cmidrule{1-5}    \multicolumn{1}{c}{\multirow{2}[2]{*}{VE}} & \multicolumn{1}{c}{$\text{MAE}_{\text{HR}}$ (bpm) $\downarrow$} & 1.76  & \textcolor[rgb]{ 0,  0,  1}{\textbf{1.68}} & \textcolor[rgb]{ 1,  0,  0}{\textbf{0.52}} \\
	
	\cmidrule{2-5}          & $\text{MAE}_{\text{RMSSD}}$ (ms) $\downarrow$ & 75.72 & \textcolor[rgb]{ 0,  0,  1}{\textbf{54.89}} & \textcolor[rgb]{ 1,  0,  0}{\textbf{12.29}} \\

	\bottomrule    \end{tabular}%
\label{tab:ablation_design2}%
\end{table}%

\begin{table}[htbp]
\fontsize{7.7}{7.7}\selectfont
\centering
\caption{Effectiveness of learnable splitting (LS) in LUs and learnable merging (LM) in ILUs.}
\begin{tabular}{ccccc}
\toprule
\multicolumn{2}{c}{Design} & w/o LS     & w/o LM     & LifWavNet \\
\midrule
\multicolumn{1}{c}{\multirow{2}[2]{*}{SR}} & \multicolumn{1}{c}{$\rho \uparrow$} & 0.566 & \textcolor[rgb]{ 0,  0,  1}{\textbf{0.707}} & \textcolor[rgb]{ 1,  0,  0}{\textbf{0.723}} \\

\cmidrule{2-5}          & \multicolumn{1}{c}{MRE $\downarrow$} & 0.639 & \textcolor[rgb]{ 0,  0,  1}{\textbf{0.468}} & \textcolor[rgb]{ 1,  0,  0}{\textbf{0.451}} \\

\cmidrule{1-5}    \multicolumn{1}{c}{\multirow{2}[2]{*}{VE}} & \multicolumn{1}{c}{$\text{MAE}_{\text{HR}}$ (bpm) $\downarrow$} & \textcolor[rgb]{ 0,  0,  1}{\textbf{2.23}} & 3.61 & \textcolor[rgb]{ 1,  0,  0}{\textbf{0.52}} \\

\cmidrule{2-5}          & $\text{MAE}_{\text{RMSSD}}$ (ms) $\downarrow$ & \textcolor[rgb]{ 0,  0,  1}{\textbf{58.65}} & 121.69 & \textcolor[rgb]{ 1,  0,  0}{\textbf{12.29}} \\

\bottomrule    \end{tabular}
\label{tab:ablation_lus1}
\vspace*{1em}

\caption{Effectiveness of CSConv (CC), self-attention (SA), and channel-attention (CA) layers in Prediction/Updation blocks.}
\begin{tabular}{cccccc}
	\toprule
	\multicolumn{2}{c}{Design} & w/o CC & w/o SA     & w/o CA     & LifWavNet \\
	\midrule
	\multicolumn{1}{c}{\multirow{2}[2]{*}{SR}} & \multicolumn{1}{c}{$\rho \uparrow$} & 0.527 & \textcolor[rgb]{ 0,  0,  1}{\textbf{0.711}} & 0.621 & \textcolor[rgb]{ 1,  0,  0}{\textbf{0.723}} \\
	
	\cmidrule{2-6}          & \multicolumn{1}{c}{MRE $\downarrow$} & 0.575 & \textcolor[rgb]{ 1,  0,  0}{\textbf{0.407}} & 0.509 & \textcolor[rgb]{ 0,  0,  1}{\textbf{0.451}} \\
	
	\midrule    \multicolumn{1}{c}{\multirow{2}[2]{*}{VE}} & \multicolumn{1}{c}{$\text{MAE}_{\text{HR}}$ (bpm) $\downarrow$} & 14.30 & 7.00 & \textcolor[rgb]{ 0,  0,  1}{\textbf{4.14}}  & \textcolor[rgb]{ 1,  0,  0}{\textbf{0.52}} \\
	
	\cmidrule{2-6}          & $\text{MAE}_{\text{RMSSD}}$ (ms) $\downarrow$ & 487.27 & 241.16 & \textcolor[rgb]{ 0,  0,  1}{\textbf{217.08}} & \textcolor[rgb]{ 1,  0,  0}{\textbf{12.29}} \\
	
	\bottomrule    \end{tabular}%
\label{tab:ablation_lus2}%
\end{table}%

\begin{table}[htbp]
\centering
\caption{Effectiveness of loss function.}
\begin{tabular}{cccccc}
\toprule
\multicolumn{2}{c}{Design} & $\mathcal{L_{T}}$     & $\mathcal{L}_1$ & $\mathcal{L}_2$     & $\mathcal{L}$ \\
\midrule
\multicolumn{1}{c}{\multirow{2}[2]{*}{SR}} & \multicolumn{1}{c}{$\rho \uparrow$} & 0.633 & \textcolor[rgb]{ 0,  0,  1}{\textbf{0.671}} & 0.661 & \textcolor[rgb]{ 1,  0,  0}{\textbf{0.723}} \\

\cmidrule{2-6}          & \multicolumn{1}{c}{MRE $\downarrow$} & 0.518 & \textcolor[rgb]{ 1,  0,  0}{\textbf{0.409}} & 0.526 & \textcolor[rgb]{ 0,  0,  1}{\textbf{0.451}} \\

\cmidrule{1-6}    \multicolumn{1}{c}{\multirow{2}[2]{*}{VE}} & \multicolumn{1}{c}{$\text{MAE}_{\text{HR}}$ (bpm) $\downarrow$} & \textcolor[rgb]{ 0,  0,  1}{\textbf{0.95}} & 1.35  & 1.52  & \textcolor[rgb]{ 1,  0,  0}{\textbf{0.52}} \\

\cmidrule{2-6}          & $\text{MAE}_{\text{RMSSD}}$ (ms) $\downarrow$ & \textcolor[rgb]{ 0,  0,  1}{\textbf{23.38}} & 71.94 & 36.54 & \textcolor[rgb]{ 1,  0,  0}{\textbf{12.29}} \\

\bottomrule    \end{tabular}
\label{tab:ablation_loss}%
\end{table}%

\noindent \textbf{Effect of parameter sharing across scales: }  We further examine whether the lifting and inverse lifting units (LUs/ILUs) should share parameters across scales. Table \ref{tab:ablation_design2} compares different network configurations (NC). In NC-I and NC-II, we share the filter parameters across scales in the analysis and synthesis stages, respectively. The results show that not sharing parameters across scales yields the best overall performance, as it allows the filters at each scale to flexibly adapt to the distinct frequency characteristics of radar to ECG mapping. Parameter sharing, while reducing complexity, limits adaptability and leads to inferior performance.

\noindent \textbf{Effectiveness of learnable splitting and merging in LU/ILU: } Table \ref{tab:ablation_lus1} evaluates the role of learnable splitting (in LUs) and learnable merging (in ILUs). Removing either mechanism significantly degrades performance, with large increases in HR and HRV estimation errors. In contrast, the full design with learnable splitting and merging achieves the best results, demonstrating that adaptive splitting and merging are critical for accurate radar‑to‑ECG reconstruction.

\noindent \textbf{Effectiveness of prediction/updation block components: } Table \ref{tab:ablation_lus2} analyzes the contribution of CSConv, self‑attention, and channel‑attention layers within the prediction and updation blocks. Removing any of these components reduces performance, with the largest drop observed when CSConv is removed. This highlights that the combination of convolutional and attention mechanisms is essential: CSConv captures local morphology, and attention mechanisms emphasize physiologically relevant features.

\noindent \textbf{Effectiveness of the loss function: } Finally, Table \ref{tab:ablation_loss} compares four different loss configurations: i) Setting $\alpha =0$ in Equation \ref{eq_loss}, we use only the temporal loss function $\mathcal{L}_T$. ii) Instead of MR-STFT, we use a single window STFT loss $\mathcal{L}_{S1}$ with window size $600$. The overall loss function is $\mathcal{L}_{1}=\mathcal{L}_{T}+0.1\mathcal{L}_{S1}$. iii) We use a single window STFT loss $\mathcal{L}_{S2}$ with window size $800$. The overall loss function is $\mathcal{L}_{2}=\mathcal{L}_{T}+0.1\mathcal{L}_{S2}$. iv) Our proposed loss function $\mathcal{L}$ in Equation \ref{eq_loss}. As the results in Table \ref{tab:ablation_loss} show, using only temporal loss $(\mathcal{L}_T)$ or single‑scale temporal–spectral similarity losses $(\mathcal{L}_{1}, \mathcal{L}_{2})$ provides partial improvements but fails to fully capture the multi‑scale nature of ECG. The proposed multi‑resolution temporal–spectral similarity loss $(\mathcal{L})$ achieves the best overall performance, confirming that joint time–frequency constraints are necessary for accurate ECG reconstruction and robust vitals estimation.

The ablation results confirm the effectiveness of LifWavNet. A four‑scale MRAS decomposition offers the best trade‑off between fine‑scale transients and low‑frequency morphology, while independent LUs/ILUs per scale adapt more flexibly than shared parameters. Learnable splitting and merging operations further enhance reconstruction and vitals estimation. Likewise, CSConv, self‑attention, and channel‑attention layers each add value by capturing the physiologically relevant features. Finally, our proposed loss function ensures a high reconstruction fidelity. Together, these components are all essential, and their integration drives LifWavNet’s superior performance.
\section{Conclusion}
\label{conclusion}
In this work, we introduced LifWavNet, a lifting wavelet-based network specifically designed for radar-to-ECG reconstruction. By formulating this reconstruction problem within a MRAS framework, our approach decomposes radar signals into scale-specific components and recombines them into physiologically meaningful ECG waveforms. LifWavNet employs learnable lifting wavelet filters, enabling the analysis and synthesis stages to adapt to an end-to-end learning for the nonlinear radar–ECG mapping.
To further enhance reconstruction accuracy, we proposed a multi-resolution STFT loss, which constrains the reconstructed ECG against the ground truth in both temporal and spectral domains, ensuring high structural fidelity. Extensive experiments on two public datasets demonstrated that LifWavNet achieves superior performance in ECG reconstruction as well as downstream HR and HRV estimation, consistently outperforming SOTA methods.
Beyond radar-to-ECG reconstruction, the proposed framework underscores the potential of learnable lifting wavelets as a general tool for cross-domain signal transformation. Future work could explore extending this paradigm to other biomedical and non-biomedical applications.

\bibliographystyle{IEEEtran}
\bibliography{main}

\begin{thebibliography}{10}
\providecommand{\url}[1]{#1}
\csname url@samestyle\endcsname
\providecommand{\newblock}{\relax}
\providecommand{\bibinfo}[2]{#2}
\providecommand{\BIBentrySTDinterwordspacing}{\spaceskip=0pt\relax}
\providecommand{\BIBentryALTinterwordstretchfactor}{4}
\providecommand{\BIBentryALTinterwordspacing}{\spaceskip=\fontdimen2\font plus
\BIBentryALTinterwordstretchfactor\fontdimen3\font minus
  \fontdimen4\font\relax}
\providecommand{\BIBforeignlanguage}[2]{{%
\expandafter\ifx\csname l@#1\endcsname\relax
\typeout{** WARNING: IEEEtran.bst: No hyphenation pattern has been}%
\typeout{** loaded for the language `#1'. Using the pattern for}%
\typeout{** the default language instead.}%
\else
\language=\csname l@#1\endcsname
\fi
#2}}
\providecommand{\BIBdecl}{\relax}
\BIBdecl

\bibitem{ecg_golden}
T.~Stracina, M.~Ronzhina, R.~Redina, and M.~Novakova, ``Golden standard or
  obsolete method? review of {ECG} applications in clinical and experimental
  context,'' \emph{Frontiers in Physiology}, vol.~13, p. 867033, 2022.

\bibitem{radar_tutorial}
G.~Paterniani, D.~Sgreccia, A.~Davoli, G.~Guerzoni, P.~Di~Viesti, A.~C.
  Valenti, M.~Vitolo, G.~M. Vitetta, and G.~Boriani, ``Radar-based monitoring
  of vital signs: A tutorial overview,'' \emph{Proceedings of the IEEE}, vol.
  111, no.~3, pp. 277--317, 2023.

\bibitem{template_tim}
Z.~Yang, H.~Yang, A.~Hu, X.~Zhuge, and J.~Miao, ``An adaptive template matching
  method for noncontact cardiac inter-beat interval detection using a 120 {GHz
  FMCW} radar,'' \emph{IEEE Transactions on Instrumentation and Measurement},
  vol.~74, pp. 1--12, 2025.

\bibitem{cardiac_tim}
W.~Xia, Y.~Li, and S.~Dong, ``Radar-based high-accuracy cardiac activity
  sensing,'' \emph{IEEE Transactions on Instrumentation and Measurement},
  vol.~70, pp. 1--13, 2021.

\bibitem{algorithms_tim}
Y.~Zhang, R.~Yang, Y.~Yue, E.~G. Lim, and Z.~Wang, ``An overview of algorithms
  for contactless cardiac feature extraction from radar signals: Advances and
  challenges,'' \emph{IEEE Transactions on Instrumentation and Measurement},
  vol.~72, pp. 1--20, 2023.

\bibitem{study_tim}
J.~Kranjec, S.~Beguš, J.~Drnovšek, and G.~Geršak, ``Novel methods for
  noncontact heart rate measurement: A feasibility study,'' \emph{IEEE
  Transactions on Instrumentation and Measurement}, vol.~63, no.~4, pp.
  838--847, 2014.

\bibitem{access_doppler}
K.~Yamamoto, R.~Hiromatsu, and T.~Ohtsuki, ``{ECG} signal reconstruction via
  doppler sensor by hybrid deep learning model with {CNN and LSTM},''
  \emph{IEEE Access}, vol.~8, pp. 130\,551--130\,560, 2020.

\bibitem{datamobilecomputing}
J.~Chen, D.~Zhang, Z.~Wu, F.~Zhou, Q.~Sun, and Y.~Chen, ``Contactless
  electrocardiogram monitoring with millimeter wave radar,'' \emph{IEEE
  Transactions on Mobile Computing}, vol.~23, no.~1, pp. 270--285, 2024.

\bibitem{rfecg}
Z.~Wang, B.~Jin, S.~Li, F.~Zhang, and W.~Zhang, ``{ECG}-grained cardiac
  monitoring using {UWB} signals,'' \emph{Proceedings of the ACM on
  Interactive, Mobile, Wearable and Ubiquitous Technologies}, vol.~6, no.~4,
  pp. 1--25, 2023.

\bibitem{radarnet}
B.~Li, W.~Li, Y.~He, W.~Zhang, and H.~Fu, ``{RadarNet}: Noncontact {ECG} signal
  measurement based on {FMCW} radar,'' \emph{IEEE Transactions on
  Instrumentation and Measurement}, vol.~73, pp. 1--9, 2024.

\bibitem{airecg}
L.~Zhao, R.~Lyu, H.~Lei, Q.~Lin, A.~Zhou, H.~Ma, J.~Wang, X.~Meng, C.~Shao,
  Y.~Tang \emph{et~al.}, ``{AirECG}: Contactless electrocardiogram for cardiac
  disease monitoring via mmwave sensing and cross-domain diffusion model,''
  \emph{Proceedings of the ACM on Interactive, Mobile, Wearable and Ubiquitous
  Technologies}, vol.~8, no.~3, pp. 1--27, 2024.

\bibitem{autoencoder_sensors}
K.-C. Liu, S.-Y. Peng, Y.~Tsao, C.-Y. Liu, Z.-A. Chen, Z.~Han~Han, W.-C. Chen,
  P.-Q. Hsieh, Y.-J. Li, Y.-J. Hsu, and S.-N. Hsu, ``A cross-modal autoencoder
  for contactless electrocardiography monitoring using frequency-modulated
  continuous wave radar,'' \emph{IEEE Sensors Journal}, vol.~24, no.~24, pp.
  41\,462--41\,473, 2024.

\bibitem{RSSRnet}
Y.~Wu, H.~Ni, C.~Mao, and J.~Han, ``Contactless reconstruction of {ECG} and
  respiration signals with mmwave radar based on {RSSRnet},'' \emph{IEEE
  Sensors Journal}, vol.~24, no.~5, pp. 6358--6368, 2024.

\bibitem{radarode}
Y.~Zhang, R.~Guan, L.~Li, R.~Yang, Y.~Yue, and E.~G. Lim, ``{radarODE}: an
  {ODE}-embedded deep learning model for contactless {ECG} reconstruction from
  millimeter-wave radar,'' \emph{IEEE Transactions on Mobile Computing}, pp.
  1--17, 2025.

\bibitem{radarodemtl}
Y.~Zhang, R.~Yang, Y.~Yue, and E.~Gee~Lim, ``{radarODE-MTL}: A multitask
  learning framework with eccentric gradient alignment for robust radar-based
  {ECG} reconstruction,'' \emph{IEEE Transactions on Instrumentation and
  Measurement}, vol.~74, pp. 1--15, 2025.

\bibitem{DCGANs}
B.~Jin, H.~Wu, Y.~Wang, Z.~Zhang, X.~Zhang, and G.~Du, ``Reconstructing
  {ECG}-like signals from millimeter-wave radar echoes: A generative
  adversarial network approach,'' \emph{IEEE Sensors Journal}, vol.~25, no.~14,
  pp. 27\,200--27\,208, 2025.

\bibitem{wavegru}
D.~Xu, Y.~Xu, K.~Xu, Z.~Hu, M.~Xing, F.~Gini, and M.~S. Greco, ``{WaveGRU-Net:
  Robust non-contact {ECG} reconstruction via MIMO millimeter-wave radar and
  multi-scale semantic analysis},'' \emph{Signal Processing}, p. 110108, 2025.

\bibitem{cardiac_model}
D.~M. Bers, ``Cardiac excitation--contraction coupling,'' \emph{Nature}, vol.
  415, no. 6868, pp. 198--205, 2002.

\bibitem{wavetheory}
S.~G. Mallat, ``A theory for multiresolution signal decomposition: the wavelet
  representation,'' \emph{IEEE transactions on pattern analysis and machine
  intelligence}, vol.~11, no.~7, pp. 674--693, 2002.

\bibitem{ecgwavelet}
P.~S. Addison, ``Wavelet transforms and the {ECG}: a review,''
  \emph{Physiological measurement}, vol.~26, no.~5, p. R155, 2005.

\bibitem{waveletdetection}
C.~Li, C.~Zheng, and C.~Tai, ``Detection of {ECG} characteristic points using
  wavelet transforms,'' \emph{IEEE Transactions on biomedical Engineering},
  vol.~42, no.~1, pp. 21--28, 1995.

\bibitem{embc_wavelet}
S.~Kundu, G.~Panda, A.~Routray, R.~Guha, and P.~Mohanty, ``Contactless
  monitoring of human vitals: A study with simultaneous measurements using
  {FMCW} radar and thermal camera,'' in \emph{2023 45th Annual International
  Conference of the IEEE Engineering in Medicine \& Biology Society (EMBC)},
  2023, pp. 1--4.

\bibitem{neurofuzzy}
T.-W. Kim and K.-C. Kwak, ``End-to-end electrocardiogram signal transformation
  from continuous-wave radar signal using deep learning model with
  maximum-overlap discrete wavelet transform and adaptive neuro-fuzzy network
  layers,'' \emph{Applied Sciences}, vol.~14, no.~19, p. 8730, 2024.

\bibitem{lifting_1996}
\BIBentryALTinterwordspacing
W.~Sweldens, ``Wavelets and the lifting scheme : A 5 minute tour,''
  \emph{Journal of Applied Mathematics and Mechanics}, vol.~76, pp. 41--44,
  1996. [Online]. Available:
  \url{https://api.semanticscholar.org/CorpusID:8440437}
\BIBentrySTDinterwordspacing

\bibitem{dawn}
M.~X.~B. Rodriguez, A.~Gruson, L.~Polania, S.~Fujieda, F.~Prieto, K.~Takayama,
  and T.~Hachisuka, ``Deep adaptive wavelet network,'' in \emph{Proceedings of
  the IEEE/CVF Winter Conference on Applications of Computer Vision}, 2020, pp.
  3111--3119.

\bibitem{compression1}
X.~Li, A.~Naman, and D.~Taubman, ``Exploration of learned lifting-based
  transform structures for fully scalable and accessible wavelet-like image
  compression,'' \emph{IEEE Transactions on Image Processing}, 2024.

\bibitem{winnet}
J.-J. Huang and P.~L. Dragotti, ``{WINNet}: Wavelet-inspired invertible network
  for image denoising,'' \emph{IEEE Transactions on Image Processing}, vol.~31,
  pp. 4377--4392, 2022.

\bibitem{wavelet_denosing}
M.~Malfait and D.~Roose, ``Wavelet-based image denoising using a markov random
  field a priori model,'' \emph{IEEE Transactions on Image Processing}, vol.~6,
  no.~4, pp. 549--565, 1997.

\bibitem{signal_classification}
\BIBentryALTinterwordspacing
M.~LIU, A.~Zeng, Q.~LAI, R.~Gao, M.~Li, J.~Qin, and Q.~Xu, ``T-wavenet: A
  tree-structured wavelet neural network for time series signal analysis,'' in
  \emph{International Conference on Learning Representations}, 2022. [Online].
  Available: \url{https://openreview.net/forum?id=U4uFaLyg7PV}
\BIBentrySTDinterwordspacing

\bibitem{1995lifting}
W.~Sweldens, ``Lifting scheme: a new philosophy in biorthogonal wavelet
  constructions,'' in \emph{Wavelet applications in signal and image processing
  III}, vol. 2569.\hskip 1em plus 0.5em minus 0.4em\relax SPIE, 1995, pp.
  68--79.

\bibitem{wavelet_hr}
Z.~Ling, W.~Zhou, Y.~Ren, J.~Wang, and L.~Guo, ``Non-contact heart rate
  monitoring based on millimeter wave radar,'' \emph{IEEE Access}, vol.~10, pp.
  74\,033--74\,044, 2022.

\bibitem{bp_neural}
Y.~Zheng, R.~Wang, and J.~Li, ``Nonlinear wavelets and bp neural networks
  adaptive lifting scheme,'' in \emph{The 2010 International Conference on
  Apperceiving Computing and Intelligence Analysis Proceeding}.\hskip 1em plus
  0.5em minus 0.4em\relax IEEE, 2010, pp. 316--319.

\bibitem{lifting_reconstruction}
G.~Piella and H.~J. Heijmans, ``Adaptive lifting schemes with perfect
  reconstruction,'' \emph{IEEE Transactions on signal processing}, vol.~50,
  no.~7, pp. 1620--1630, 2002.

\bibitem{channel_attention}
Q.~Wang, B.~Wu, P.~Zhu, P.~Li, W.~Zuo, and Q.~Hu, ``{ECA-Net}: Efficient
  channel attention for deep convolutional neural networks,'' in
  \emph{Proceedings of the IEEE/CVF conference on computer vision and pattern
  recognition}, 2020, pp. 11\,534--11\,542.

\bibitem{latis}
G.~Panda, S.~Kundu, S.~Bhattacharya, and A.~Routray, ``{LATIS}: Lambda
  abstraction-based thermal image super-resolution,'' \emph{arXiv preprint
  arXiv:2311.12046}, 2023.

\bibitem{layerNorm}
J.~Xu, X.~Sun, Z.~Zhang, G.~Zhao, and J.~Lin, ``Understanding and improving
  layer normalization,'' \emph{Advances in Neural Information Processing
  Systems}, vol.~32, 2019.

\bibitem{vit}
A.~Dosovitskiy, L.~Beyer, A.~Kolesnikov, D.~Weissenborn, X.~Zhai,
  T.~Unterthiner, M.~Dehghani, M.~Minderer, G.~Heigold, S.~Gelly, J.~Uszkoreit,
  and N.~Houlsby, ``An image is worth 16x16 words: Transformers for image
  recognition at scale,'' in \emph{ICLR}, 2021.

\bibitem{nature}
S.~Schellenberger, K.~Shi, T.~Steigleder, A.~Malessa, F.~Michler, L.~Hameyer,
  N.~Neumann, F.~Lurz, R.~Weigel, C.~Ostgathe \emph{et~al.}, ``A dataset of
  clinically recorded radar vital signs with synchronised reference sensor
  signals,'' \emph{Scientific data}, vol.~7, no.~1, p. 291, 2020.

\bibitem{elsevier}
K.~Edanami and G.~Sun, ``Medical radar signal dataset for non-contact
  respiration and heart rate measurement,'' \emph{Data in brief}, vol.~40, p.
  107724, 2022.

\bibitem{neonatal}
G.~Beltr{\~a}o, R.~Stutz, F.~Hornberger, W.~A. Martins, D.~Tatarinov,
  M.~Alaee-Kerahroodi, U.~Lindner, L.~Stock, E.~Kaiser, S.~Goedicke-Fritz
  \emph{et~al.}, ``Contactless radar-based breathing monitoring of premature
  infants in the neonatal intensive care unit,'' \emph{Scientific Reports},
  vol.~12, no.~1, p. 5150, 2022.

\end{thebibliography}

\end{document}